\documentclass[10pt,twocolumn,letterpaper]{article}

\usepackage{iccv}
\usepackage{times}
\usepackage{epsfig}
\usepackage{graphicx}
\usepackage{amsmath}
\usepackage{amssymb}
\usepackage{mathtools}
\usepackage{multirow}
\usepackage{enumitem}
\usepackage{caption}
\usepackage{subcaption}
\usepackage{array}
\usepackage{colortbl}
\usepackage{booktabs}
\usepackage{bbm}
\usepackage{multicol}
\usepackage{makecell}
\usepackage{xcolor}
\usepackage{xspace}
\usepackage{float}
\usepackage{color}
\usepackage{pifont}
\usepackage{marvosym}
\setitemize[1]{itemsep=0pt,partopsep=0pt,parsep=\parskip,topsep=5pt}
\newcommand{\tablestyle}[2]{\setlength{\tabcolsep}{#1}\renewcommand{\arraystretch}{#2}\centering\footnotesize}

\usepackage[pagebackref=true,breaklinks=true,letterpaper=true,colorlinks,bookmarks=false]{hyperref}

\iccvfinalcopy 

\def\iccvPaperID{2578} 

\ificcvfinal\fi

\newlength\savedwidth
\newcommand\whline{\noalign{\global\savedwidth\arrayrulewidth\global\arrayrulewidth 0.8pt}\hline\noalign{\global\arrayrulewidth\savedwidth}}

\newcommand{\green}[1]{\textcolor[RGB]{96,177,87}{#1}}
\newcommand{\fn}[1]{\footnotesize{#1}}
\newcommand{\gbf}[1]{\green{\bf{\fn{(#1)}}}}

\newcommand{\tabincell}[2]{\begin{tabular}{@{}#1@{}}#2\end{tabular}}
\definecolor{mygray}{gray}{.92}
	
\newcommand\blfootnote[1]{%
\begingroup
\renewcommand\thefootnote{}\footnote{#1}%
\addtocounter{footnote}{-1}%
\endgroup
}
	
\def\ie{\emph{i.e.}}
\def\eg{\emph{e.g.}}
\def\etc{\emph{etc}}
\def\etal{{\em et al.~}}

\begin{document}

\title{Pyramid Vision Transformer: A Versatile Backbone for Dense Prediction without Convolutions}

\author{
    Wenhai Wang$^{1}$, 
    Enze Xie$^{2}$,
    Xiang Li$^{3}$,
    Deng-Ping Fan$^{4}$\textsuperscript{\Letter}, \\
    Kaitao Song$^{3}$,
    Ding Liang$^{5}$,
    Tong Lu$^{1}$\textsuperscript{\Letter},
    Ping Luo$^{2}$,
    Ling Shao$^{4}$\\
    $^1$Nanjing University~~~
    $^2$The University of Hong Kong~~~\\
    $^3$Nanjing University of Science and Technology~~~
    $^4$IIAI~~~
    $^5$SenseTime Research\\
    {\small \url{https://github.com/whai362/PVT}}
}

\ificcvfinal\thispagestyle{empty}\fi

\twocolumn[{%
\maketitle
\vspace{-10mm}
\begin{figure}[H]
\hsize=\textwidth
\centering
\hspace{-0.2in}
\begin{subfigure}{0.30\textwidth}
    \centering
    \includegraphics[width=0.95\textwidth]{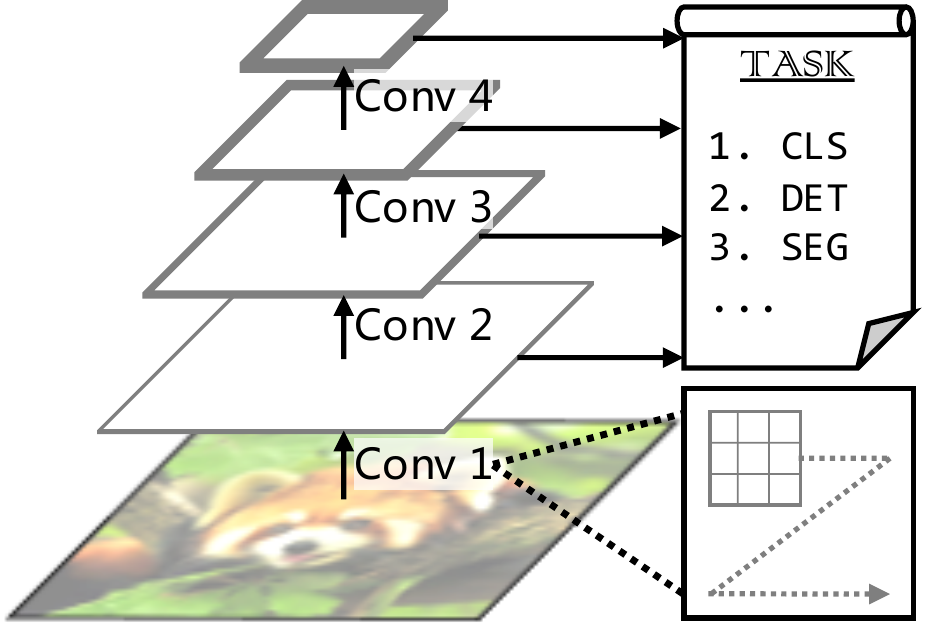}
    \caption{CNNs: VGG~\cite{simonyan2014very}, ResNet~\cite{he2016deep}, \etc.}
    \label{fig:1a}	
\end{subfigure}    
\hspace{0.2in}
\begin{subfigure}{0.30\textwidth}
     \centering
     \includegraphics[width=0.95\textwidth]{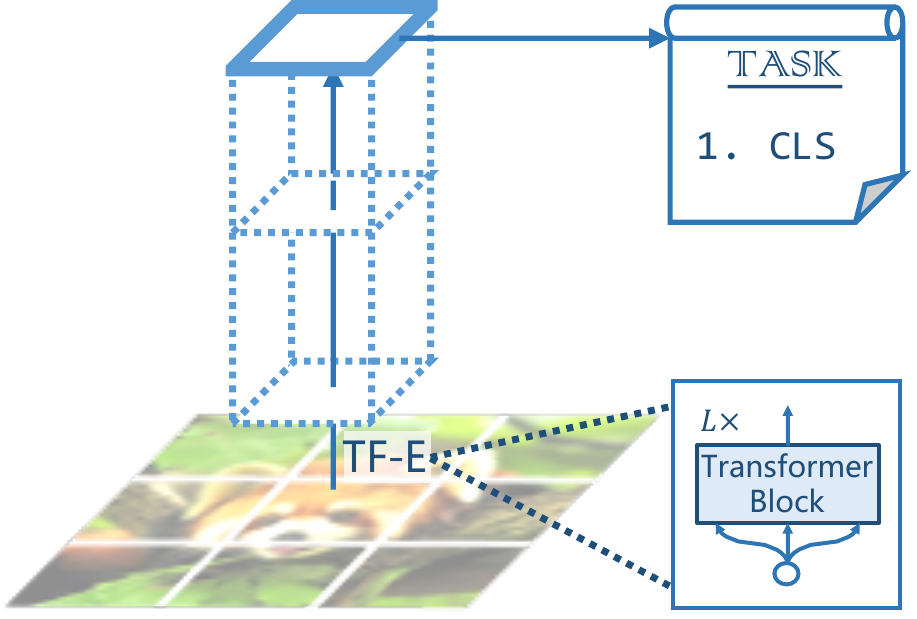}
     \caption{Vision Transformer~\cite{dosovitskiy2020image}}
     \label{fig:1b}
\end{subfigure}
\hspace{0.2in}
\begin{subfigure}{0.30\textwidth}
     \centering
     \includegraphics[width=0.95\textwidth]{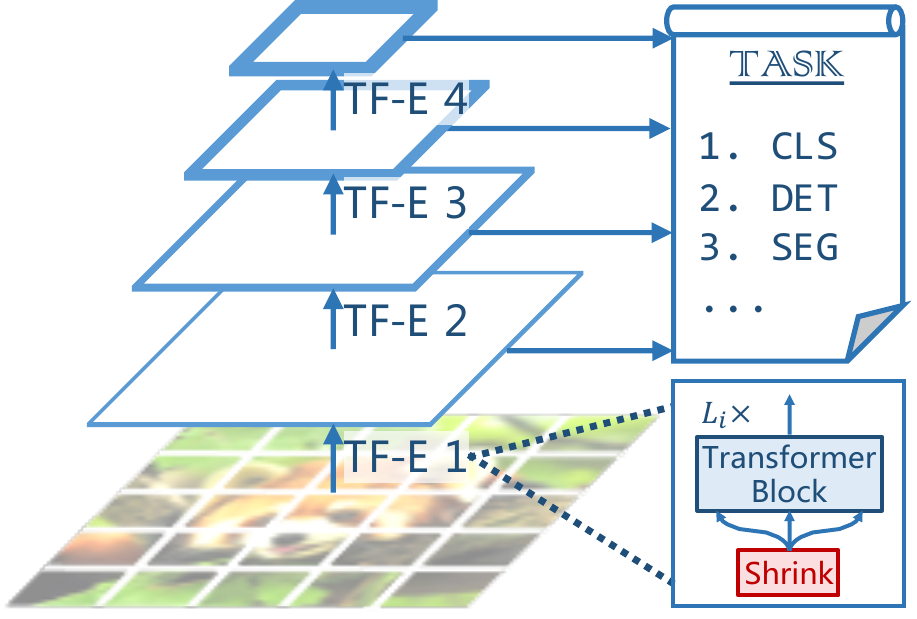}
     \caption{Pyramid Vision Transformer (ours)}
     \label{fig:1d}
\end{subfigure}
\vspace{-3mm}
\caption{\textbf{Comparisons of different architectures,} where ``Conv'' and ``TF-E''
stand for ``convolution'' and ``Transformer encoder'', 
respectively. 
(a) Many CNN backbones use a pyramid structure for dense prediction tasks such as object detection (DET), instance and semantic segmentation (SEG).
(b) The recently proposed Vision Transformer (ViT)~\cite{dosovitskiy2020image} is a ``columnar'' structure specifically designed for image classification (CLS). (c) By incorporating the pyramid structure from CNNs, we present the Pyramid Vision Transformer (PVT), which can be used as a versatile backbone for many computer vision tasks,
broadening the scope and impact of ViT.
Moreover, our experiments also show that PVT can easily be combined with DETR~\cite{carion2020end} to build an end-to-end object detection system without convolutions.
}
\label{fig:pipeline}
\end{figure}
\vspace{2mm}
}]

\begin{abstract}
 
Although convolutional neural networks (CNNs) have achieved great success in computer vision, this work investigates a simpler, convolution-free backbone network useful for many dense prediction tasks.
Unlike the recently-proposed Vision Transformer (ViT) that was designed for image classification specifically, we introduce the Pyramid Vision Transformer~(PVT)\blfootnote{\Letter~Corresponding authors: Deng-Ping Fan (dengpfan@gmail.com); Tong Lu (lutong@nju.edu.cn).}, which overcomes the difficulties of porting Transformer to various dense prediction tasks. PVT has several merits compared to current state
of the arts.
(1) Different from ViT that typically yields low-resolution outputs and incurs high computational and memory costs, PVT not only can be trained on dense partitions of an image to achieve high output resolution,
which is important for dense prediction, but also uses a progressive shrinking pyramid to reduce the computations of large feature maps.
(2) PVT inherits the advantages of both CNN and Transformer, making it a unified backbone for various vision tasks without convolutions, where it can be used as a direct replacement for CNN backbones. 
(3) We validate PVT through extensive experiments, showing that it boosts the performance of many downstream tasks, including object detection, instance and semantic segmentation.
For example, with a comparable number of parameters, PVT+RetinaNet achieves 40.4 AP on the COCO dataset, surpassing ResNet50+RetinNet (36.3 AP) by 4.1 absolute AP (see Figure \ref{fig:cmp}).
We hope that PVT could serve as an alternative and useful backbone for pixel-level predictions and facilitate future research.

\end{abstract}

 
\section{Introduction}

Convolutional neural network (CNNs) have achieved remarkable success in  computer vision, making them a versatile and dominant approach for almost all tasks~
\cite{simonyan2014very,he2016deep,xie2017aggregated,ren2015faster,he2017mask,lin2017focal,chen2018encoder,kirillov2019panoptic}. 
Nevertheless, this work aims to explore an alternative backbone network beyond CNN,
which can be used for dense prediction tasks such as object detection~\cite{lin2014microsoft,everingham2010pascal}, semantic~\cite{zhou2017scene} and instance segmentation~\cite{lin2014microsoft}, in addition to image classification~\cite{deng2009imagenet}.

\begin{figure}[t]
	\centering
	\includegraphics[width=0.98\columnwidth]{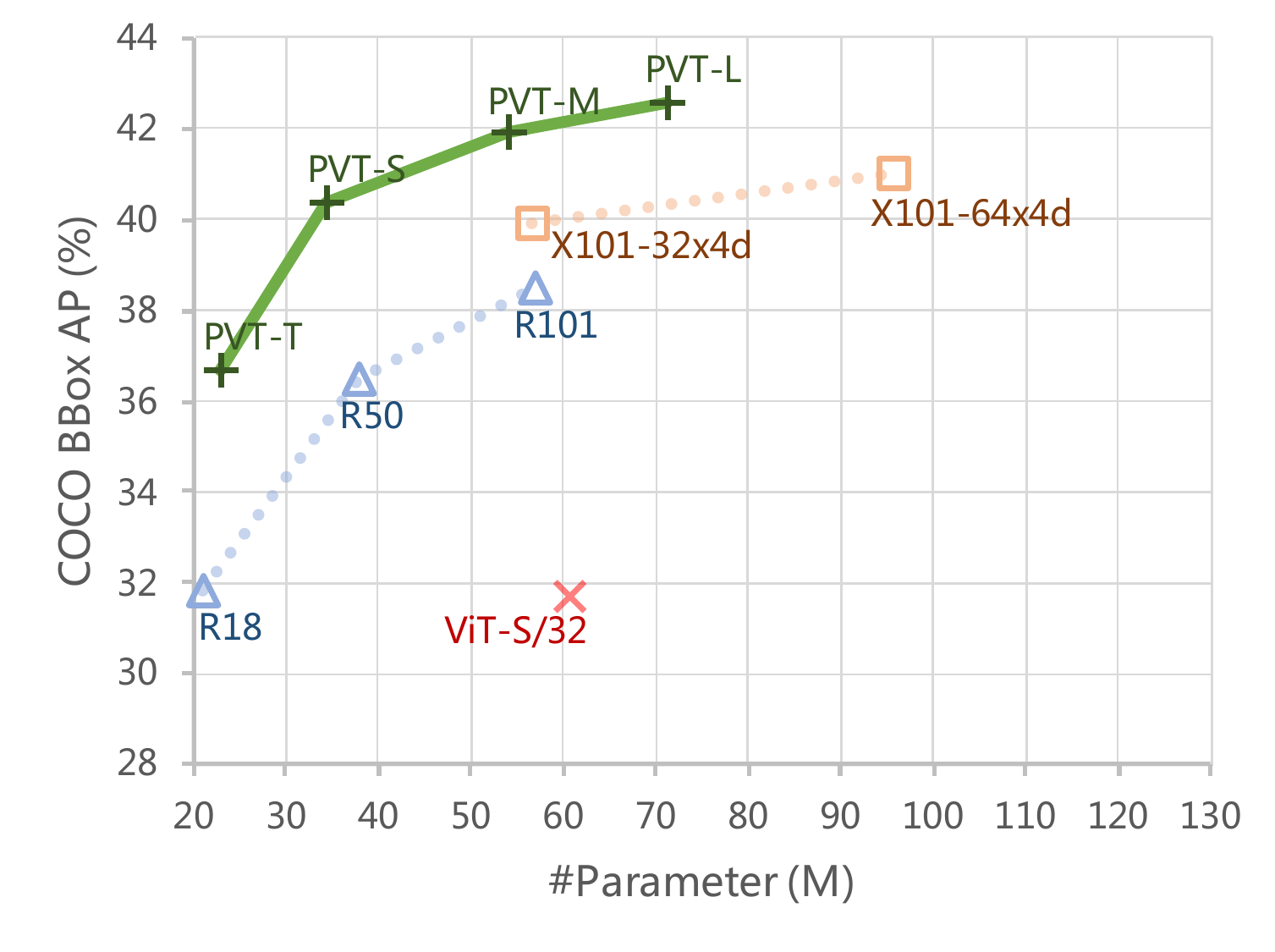}
	\hspace{-42.5mm}\resizebox{.48\columnwidth}{!}{\tablestyle{2pt}{1}
	\begin{tabular}[b]{l|c|c}
	\renewcommand{\arraystretch}{0.1}
	Backbone & \#Param (M) & AP \\
	\whline
	R18~\cite{he2016deep} & 21.3 & 31.8 \\
	\rowcolor{mygray}
	PVT-T (ours) & 23.0 & 36.7 \\
	\hline
	R50~\cite{he2016deep} & 37.7 & 36.3 \\
	\rowcolor{mygray}
	PVT-S (ours) & 34.2 & 40.4 \\
	\hline
	R101~\cite{he2016deep} & 56.7 & 38.5 \\
	X101-32x4d~\cite{xie2017aggregated} & 56.4 & 39.9 \\
	ViT-S/32~\cite{dosovitskiy2020image} & 60.8  & 31.7 \\
	\rowcolor{mygray}
	PVT-M (ours) & 53.9 & 41.9 \\
	\hline
	X101-64x4d~\cite{xie2017aggregated} & 95.5 & 41.0 \\
	\rowcolor{mygray}
	PVT-L (ours) & 71.1 & 42.6 \\
			\multicolumn{3}{l}{
			\vspace{8.3mm}}\\
\end{tabular}
	}
    
   \caption{\textbf{Performance comparison on COCO \texttt{val2017} of different backbones using RetinaNet for object detection}, where ``T'', ``S'', ``M'' and ``L'' denote our PVT models with tiny, small, medium and large size. We see that when the number of parameters among different models are comparable, PVT variants significantly outperform their corresponding counterparts such as ResNets (R)~\cite{he2016deep}, ResNeXts (X)~\cite{xie2017aggregated}, and ViT \cite{dosovitskiy2020image}.
}

\label{fig:cmp}
\end{figure}

Inspired by the success of Transformer~\cite{vaswani2017attention} in natural language processing,
many researchers have explored its application in computer vision.
For example, some works~\cite{carion2020end,zhu2020deformable,xie2021segmenting,sun2020transtrack,hu2021Transformer,liu2021vst} model the vision task as a dictionary lookup problem with learnable queries, and use the Transformer decoder
as a task-specific head on top of the CNN backbone.
Although some prior arts have also incorporated attention modules~\cite{wang2018non,ramachandran2019stand,zhao2020exploring} into CNNs, 
as far as we know, \emph{exploring a clean and convolution-free Transformer backbone to address dense prediction tasks in computer vision is rarely studied}.

Recently, Dosovitskiy \etal\cite{dosovitskiy2020image} introduced the Vision Transformer (ViT) for image classification.
This is an interesting and meaningful attempt to replace the CNN backbone with a convolution-free model. 
As shown in Figure~\ref{fig:pipeline} (b), ViT has a columnar structure with coarse image patches
as input.\footnote{Due to resource constraints, ViT cannot use fine-grained image patches (\eg, 4$\times$4 pixels per patch) as input, instead only receive coarse patches (\eg, 32$\times$32 pixels per patch) as input, which leads to its low output resolution (\eg, 32-stride).}
Although ViT is applicable to image classification, it is challenging to directly 
adapt it to pixel-level dense predictions such as object detection and segmentation, because 
(1) its output feature map is single-scale and low-resolution,
and (2) its computational and memory costs are relatively high even for common input image sizes (\eg, shorter edge of 800 pixels in the COCO benchmark~\cite{lin2014microsoft}).

To address the above limitations, 
this work proposes a pure Transformer backbone,
termed Pyramid Vision Transformer (PVT), which can serve as an alternative to the CNN backbone in many 
downstream tasks, including image-level prediction as well as  pixel-level dense predictions.
Specifically, as illustrated in Figure~\ref{fig:pipeline} (c), 
our PVT overcomes the difficulties of the conventional Transformer by (1) taking fine-grained image patches (\ie, 4$\times$4 pixels per patch) as input to learn high-resolution representation, which is essential for dense prediction tasks;
(2) introducing a progressive shrinking pyramid to reduce the sequence length of Transformer as the network deepens,
significantly reducing the computational cost, and (3) adopting a spatial-reduction attention (SRA) layer to further reduce the resource consumption when learning high-resolution features.

Overall, the proposed PVT possesses the following merits.
Firstly, compared to the 
traditional
CNN backbones (see Figure~\ref{fig:pipeline} (a)), which have local receptive fields that increase with the network depth, our PVT always produces a global receptive field,
which is more suitable for detection and segmentation.
Secondly, compared to ViT (see Figure~\ref{fig:pipeline} (b)), thanks to its advanced pyramid structure, our method can more easily be plugged into many representative dense prediction pipelines, \eg, RetinaNet~\cite{lin2017focal} and Mask R-CNN~\cite{he2017mask}.
Thirdly,
we can build a convolution-free pipeline by combining our PVT with other task-specific
Transformer decoders,
such as PVT+DETR~\cite{carion2020end} for object detection.
\emph{To our knowledge, this is the first entirely convolution-free object detection pipeline.}

Our main contributions are as follows:

    (1) We propose Pyramid Vision Transformer (PVT), which is the first pure Transformer backbone designed for various pixel-level dense prediction tasks.
    Combining our PVT and DETR, we can construct an end-to-end object detection system without convolutions and handcrafted components such as dense anchors and non-maximum suppression (NMS).
    
    (2) We overcome many difficulties when porting Transformer to dense
    predictions, by designing a
    progressive shrinking pyramid and a spatial-reduction attention (SRA). These are able to reduce the resource consumption of Transformer, making PVT flexible to learning multi-scale and high-resolution features.
    
    (3) We evaluate the proposed PVT on several different tasks, including image classification, object detection, instance and semantic segmentation, and compare it with 
    popular
    ResNets~\cite{he2016deep} and ResNeXts~\cite{xie2017aggregated}. As presented in Figure \ref{fig:cmp}, 
    our PVT with different parameter scales
    can consistently archived improved performance compared to the prior arts.
    For example, under a comparable number of parameters, using RetinaNet~\cite{lin2017focal} for object detection, PVT-Small achieves 40.4 AP on COCO \texttt{val2017}, outperforming ResNet50 by 4.1 points (40.4 \vs 36.3).
    Moreover, PVT-Large achieves 42.6 AP, which is 1.6 points better than ResNeXt101-64x4d, with  30\% less parameters.

\section{Related Work}

\subsection{CNN Backbones}

CNNs are the work-horses of deep neural networks in visual recognition. The standard
CNN was first introduced in \cite{lecun1998gradient} to distinguish handwritten numbers. 
The model
contains convolutional kernels with a certain receptive field that captures favorable visual context. To provide translation equivariance, the weights of convolutional kernels are shared over the entire image space. More recently, with the rapid development of the computational resources (\eg, GPU), the successful training of stacked convolutional blocks~\cite{krizhevsky2012imagenet,simonyan2014very}
on large-scale image classification datasets (\eg, ImageNet \cite{russakovsky2015imagenet}) has become possible. For instance, GoogLeNet \cite{szegedy2015going} demonstrated that a convolutional operator containing multiple kernel paths can achieve very competitive performance. 
The effectiveness of a multi-path convolutional block was further validated in Inception series \cite{szegedy2016rethinking,szegedy2017inception}, ResNeXt \cite{xie2017aggregated}, DPN \cite{chen2017dual}, MixNet \cite{wang2018mixed} and SKNet \cite{li2019selective}.
Further, ResNet \cite{he2016deep} introduced skip connections into the convolutional block, making it possible to create/train very deep networks and obtaining impressive results in the field of computer vision.
DenseNet \cite{huang2017densely} introduced a densely connected topology, which connects each convolutional block to all previous blocks. 
More recent advances can be found in recent survey/review papers~\cite{khan2020survey,shorten2019survey}.

Unlike the full-blown CNNs, the vision Transformer backbone is still in its early stage of development. In this work, we try to extend the scope of Vision Transformer by designing a new versatile Transformer backbone suitable for most vision tasks.

\subsection{Dense Prediction Tasks}

\textbf{Preliminary.} The dense prediction task aims to perform pixel-level classification or regression on a feature map.
Object detection and semantic segmentation are two representative dense prediction tasks.

\textbf{Object Detection.} In the era of deep learning, CNNs~\cite{lecun1998gradient} have become the dominant framework for object detection, which includes single-stage detectors (\eg, SSD~\cite{liu2016ssd}, RetinaNet \cite{lin2017focal}, FCOS \cite{tian2019fcos}, GFL \cite{li2020generalized,li2020generalizedv2}, PolarMask \cite{xie2020polarmask} and OneNet \cite{sun2020onenet}) and multi-stage detectors (Faster R-CNN \cite{ren2015faster}, Mask R-CNN \cite{he2017mask}, Cascade R-CNN \cite{cai2018cascade} and Sparse R-CNN \cite{sun2020sparse}).
Most of these popular object detectors are built on high-resolution or multi-scale feature maps to obtain good detection performance.
Recently, DETR~\cite{carion2020end} and deformable DETR~\cite{zhu2020deformable} combined the CNN backbone and the Transformer decoder to build an end-to-end object detector.
Likewise, they also require high-resolution or multi-scale feature maps for accurate object detection.

\textbf{Semantic Segmentation.} CNNs also play an important role in semantic segmentation.
In the early stages, FCN \cite{long2015fully} introduced a fully convolutional architecture to generate a spatial segmentation map for a given image of any size.
After that, the deconvolution operation was introduced by Noh \etal\cite{noh2015learning} and achieved impressive performance on the PASCAL VOC 2012 dataset \cite{shetty2016application}.
Inspired by FCN, U-Net \cite{ronneberger2015u} was proposed for the medical image segmentation domain specifically, bridging the information flow between corresponding low-level and high-level feature maps of the same spatial sizes. 
To explore richer global context representation, Zhao \etal\cite{zhao2017pyramid} designed a pyramid pooling module over various pooling scales, and Kirillov \etal\cite{kirillov2019panoptic} developed a lightweight segmentation head termed Semantic FPN, based on FPN~\cite{lin2017feature}.
Finally, the DeepLab family \cite{chen2017deeplab,liu2019auto} applies dilated convolutions to enlarge the receptive field while maintaining the feature map resolution.
Similar to object detection methods, semantic segmentation models also rely on high-resolution or multi-scale feature maps.

\begin{figure*}[t]
		\centering
		\setlength{\fboxrule}{0pt}
		\fbox{\includegraphics[width=0.95\textwidth]{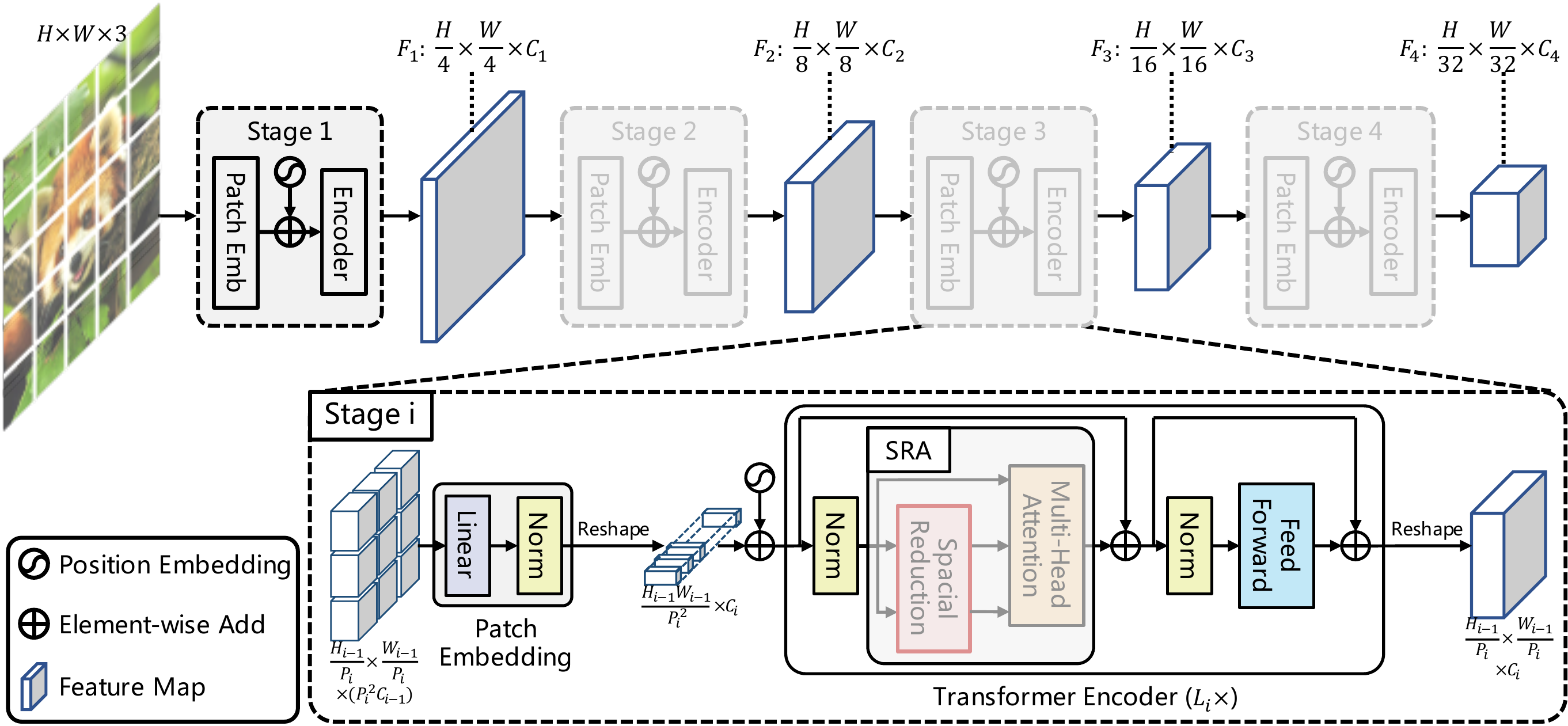}}
		\caption{\textbf{Overall architecture of Pyramid Vision Transformer (PVT).} The entire model is divided into four stages, each of which is comprised of a patch embedding layer and a $L_i$-layer Transformer encoder. Following a pyramid structure, the output resolution of the four stages progressively shrinks from high (4-stride) to low (32-stride).}
		\label{fig:arch}
\end{figure*}

\subsection{Self-Attention and Transformer in Vision}

As convolutional filter weights are usually fixed after training, 
they cannot be dynamically adapted to different inputs.
Many methods have been proposed to alleviate this problem using dynamic filters~\cite{jia2016dynamic} or self-attention operations~\cite{vaswani2017attention}.
The non-local block \cite{wang2018non} attempts to model long-range dependencies in both space and time, which has been shown beneficial for accurate video classification.
However, despite its success, the non-local operator suffers from the high computational and memory costs. Criss-cross \cite{huang2019ccnet} further reduces the complexity by generating sparse attention maps through a criss-cross path. 
Ramachandran \etal\cite{ramachandran2019stand} proposed the stand-alone self-attention to replace convolutional layers with local self-attention units. 
AANet \cite{bello2019attention} achieves competitive results when combining the self-attention and convolutional operations.
LambdaNetworks~\cite{bello2021lambdanetworks} uses the lambda layer, an efficient self-attention to replace the convolution in the CNN.
DETR \cite{carion2020end} utilizes the Transformer decoder to model object detection as an end-to-end dictionary lookup problem with learnable queries,
successfully removing the need for handcrafted processes such as NMS. 
Based on DETR, deformable DETR \cite{zhu2020deformable} further adopts a deformable attention layer to focus on a sparse set of contextual elements, obtaining faster convergence and better performance.
Recently, Vision Transformer (ViT) \cite{dosovitskiy2020image} employs a pure Transformer~\cite{vaswani2017attention} model for image classification by treating an image as a sequence of patches.
DeiT \cite{touvron2020training} further extends ViT using a novel distillation approach. 
Different from previous models, this work introduces the pyramid structure into Transformer to present a pure Transformer backbone for dense prediction tasks, rather than a task-specific head or an image classification model.

\section{Pyramid Vision Transformer (PVT)}

\subsection{Overall Architecture}

Our goal is to introduce the pyramid structure into the Transformer framework, so that it can generate multi-scale feature maps for dense prediction tasks (\eg, object detection and semantic segmentation).
An overview of PVT is depicted in Figure~\ref{fig:arch}.
Similar to CNN backbones~\cite{he2016deep}, our method has four stages that generate feature maps of different scales.
All stages share a similar architecture, which consists of a patch embedding layer and $L_i$ Transformer encoder layers.

In the first stage, given an input image of size $H\!\times\!W\! \times\!3$, we first divide it into $\frac{HW}{4^2}$ patches,\footnote{As done for ResNet, we keep the highest  resolution of our output feature map at 4-stride.} each of size $4\!\times\!4\!\times\!3$.
Then, we feed the flattened patches to a linear projection and obtain embedded patches of size $\frac{HW}{4^2}\!\times\!C_1$.
After that, the embedded patches along with a position embedding  are passed through a Transformer encoder with $L_1$ layers, and the output is reshaped to a feature map $F_1$ of size $\frac{H}{4}\!\times\!\frac{W}{4}\!\times\!C_1$. 
In the same way, using the feature map from the previous stage as input, we obtain the following feature maps: $F_2$, $F_3$, and $F_4$, whose strides are 8, 16, and 32 pixels with respect to the input image.
With the feature pyramid $\{F_1, F_2, F_3, F_4\}$, our method can be easily applied to most downstream tasks, including image classification, object detection, and semantic segmentation.

\subsection{Feature Pyramid for Transformer}

Unlike CNN backbone networks~\cite{simonyan2014very,he2016deep}, which use different convolutional strides to obtain multi-scale feature maps, our PVT uses a \textit{progressive shrinking strategy} to control the scale of feature maps by patch embedding layers.

Here, we denote the patch size of the $i$-th stage as $P_i$.
At the beginning of stage $i$, we first evenly divide the input feature map $F_{i\!-\!1}\!\in\! \mathbb{R}^{H_{i\!-\!1}\!\times\!W_{i\!-\!1}\!\times\! C_{i\!-\!1}}$ into $\frac{H_{i\!-\!1}W_{i\!-\!1}}{P_i^2}$ patches, 
and then each patch is flatten and projected to a $C_i$-dimensional embedding. 
After the linear projection, the shape of the embedded patches can be viewed as $\frac{H_{i\!-\!1}}{P_i}\!\times\!\frac{W_{i\!-\!1}}{P_i}\!\times\!C_i$, where the height and width are $P_i$ times smaller than the input.

In this way, we can flexibly adjust the scale of the feature map in each stage, making it possible to construct a feature pyramid for Transformer.

\subsection{Transformer Encoder}

\begin{figure}
		\centering
		\setlength{\fboxrule}{0pt}
		\fbox{\includegraphics[width=0.4\textwidth]{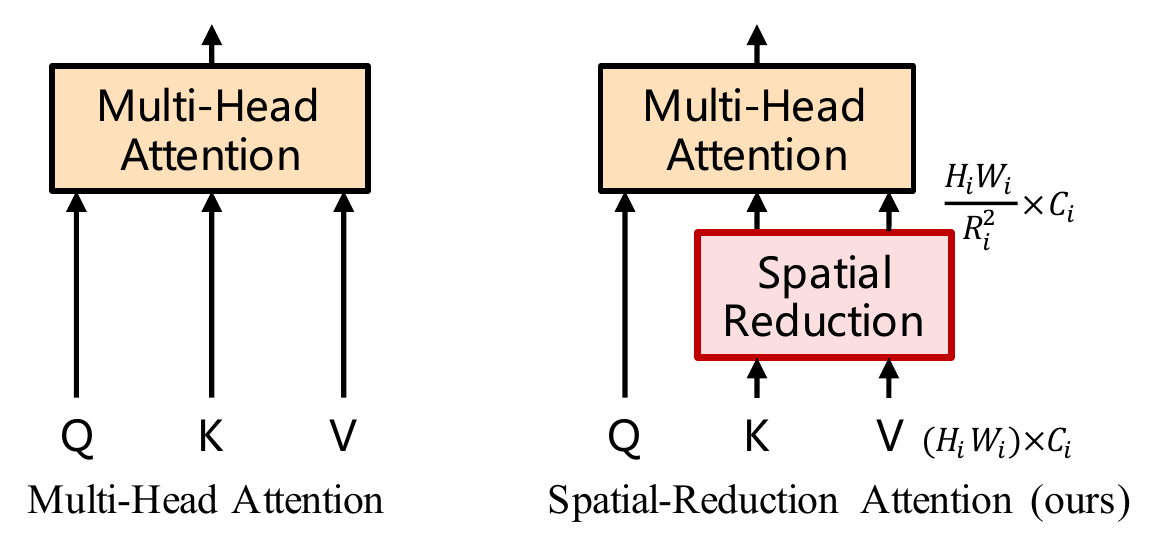}}
		 
		\caption{\textbf{Multi-head attention (MHA) \vs spatial-reduction attention (SRA).} 
		With the spatial-reduction operation, the computational/memory cost of our SRA is much lower than that of MHA.
		}
		\label{fig:att}
\end{figure}

The Transformer encoder in the stage $i $ has $L_i$ encoder layers, each of which is composed of an attention layer and a feed-forward layer~\cite{vaswani2017attention}.
Since PVT needs to process high-resolution (\eg, 4-stride) feature maps, 
we propose a spatial-reduction attention (SRA) layer to replace the traditional multi-head attention (MHA) layer~\cite{vaswani2017attention} in the encoder.

Similar to MHA, our SRA receives a query $Q$, a key $K$, and a value $V$ as input, and outputs a refined feature.
The difference is that our SRA reduces the spatial scale of $K$ and $V$ before the attention operation (see Figure~\ref{fig:att}), which largely reduces the computational/memory overhead.
Details of the SRA in the stage $i$ can be formulated as follows:
\begin{equation}
    {\rm SRA}(Q, K, V) = {\rm Concat}({\rm head}_0,... , {\rm head}_{N_i})W^O,
    \label{eqn:att1}
\end{equation}
\begin{equation}
    {\rm head}_j={\rm Attention}(QW_j^Q, {\rm SR}(\!K\!)W_j^K, {\rm SR}(\!V\!)W_j^V),
    \label{eqn:att2}
\end{equation}
where ${\rm Concat}(\cdot)$ is the concatenation operation as in \cite{vaswani2017attention}. $W_j^Q\!\in\!\mathbb{R}^{C_i\!\times\!d_{\rm head}}$,
$W_j^K\!\in\!\mathbb{R}^{C_i\!\times\!d_{\rm head}}$,
$W_j^V\!\in\!\mathbb{R}^{C_i\!\times\!d_{\rm head}}$, and
$W^O\!\in\!\mathbb{R}^{C_i\!\times\!C_i}$
are linear projection  parameters. 
$N_i$ is the head number of the attention layer in Stage $i$. 
Therefore, the dimension of each head (\ie, $d_{\rm head}$) is equal to $\frac{C_i}{N_i}$.
${\rm SR}(\cdot)$ is the operation for reducing the spatial dimension of the input sequence (\ie, $K$ or $V$), which is written as:
 
\begin{equation}
    {\rm SR}(\mathbf{x}) = {\rm Norm}({\rm Reshape}(\mathbf{x}, R_i)W^S).
    \label{eqn:att4}
\end{equation}
Here, $\mathbf{x}\!\in\!\mathbb{R}^{(H_iW_i)\!\times\!C_i}$ represents a input sequence, and $R_i$ denotes the reduction ratio of the attention layers in Stage $i$.
${\rm Reshape}(\mathbf{x}, R_i)$ is an operation of reshaping the input sequence $\mathbf{x}$
to a sequence of size $\frac{H_iW_i}{R_i^2}\!\times\!(R_i^2C_i)$.
$W_S\!\in\!\mathbb{R}^{(R_i^2C_i)\!\times\!C_i}$
is a linear projection that reduces the dimension of the input sequence to $C_i$. ${\rm Norm}(\cdot)$ refers to layer normalization~\cite{ba2016layer}.
As in the original Transformer~\cite{vaswani2017attention},
our attention operation ${\rm Attention}(\cdot)$ is calculated as:
\begin{equation}
    {\rm Attention}(\mathbf{q}, \mathbf{k}, \mathbf{v}) = {\rm Softmax}(\frac{\mathbf{q}\mathbf{k}^\mathsf{T}}{\sqrt{d_{\rm head}}})\mathbf{v}.
    \label{eqn:att3}
\end{equation}
Through these formulas, we can find that the computational/memory costs of our attention operation are $R_i^2$ times lower than those of MHA, so our SRA can handle larger input feature maps/sequences with limited resources.

\subsection{Model Details}

In summary, the hyper parameters of our method are listed as follows:
\begin{itemize}
    \item $P_i$: the patch size of Stage $i$;
    \item $C_i$: the channel number of the output of Stage $i$;
    \item $L_i$: the number of encoder layers in Stage $i$;
    \item $R_i$: the reduction ratio of the SRA in Stage $i$;
    \item $N_i$: the head number of the SRA in Stage $i$;
    \item $E_i$: the expansion ratio of the feed-forward layer~\cite{vaswani2017attention} in Stage $i$;
\end{itemize}
Following the design rules of ResNet~\cite{he2016deep}, we (1) use small output channel numbers in shallow stages; and (2) concentrate the major computation resource
in intermediate stages.

To provide instances for discussion, we describe a series of PVT models with different scales, namely PVT-Tiny, -Small, -Medium, and -Large, in Table~\ref{tab:arch}, whose parameter numbers are comparable to ResNet18, 50, 101, and 152 respectively.
More details of employing these models in specific downstream tasks will be introduced in Section~\ref{sec:task}.
\begin{table*}[t]
    \centering
    \renewcommand\arraystretch{ 1.0}
    \setlength{\tabcolsep}{1.0mm}
    \begin{tabular}{c|c|c|c|c|c|c}
    \renewcommand{\arraystretch}{0.1}
	 & Output Size & Layer Name & PVT-Tiny & PVT-Small & PVT-Medium  & PVT-Large \\
	\whline
	\multirow{2}{*}[-2.5ex]{Stage 1} & \multirow{2}{*}[-2.5ex]{\scalebox{1.3}{$\frac{H}{4}\times \frac{W}{4}$}} & Patch Embedding & \multicolumn{4}{c}{$P_1=4$;\ \ \ $C_1=64$} \\
	\cline{3-7}
	& & \tabincell{c}{Transformer\\Encoder} & 
	$\begin{bmatrix}
	\begin{array}{l}
	R_1=8 \\
	N_1=1 \\
	E_1=8 \\
	\end{array}
	\end{bmatrix} \times 2$ &
	$\begin{bmatrix}
	\begin{array}{l}
	R_1=8 \\
	N_1=1 \\
	E_1=8 \\
	\end{array}
	\end{bmatrix} \times 3$ &
	$\begin{bmatrix}
	\begin{array}{l}
	R_1=8 \\
	N_1=1 \\
	E_1=8 \\
	\end{array}
	\end{bmatrix} \times 3$ &
	$\begin{bmatrix}
	\begin{array}{l}
	R_1=8 \\
	N_1=1 \\
	E_1=8 \\
	\end{array}
	\end{bmatrix} \times 3$ \\
	\hline
	\multirow{2}{*}[-2.5ex]{Stage 2} & \multirow{2}{*}[-2.5ex]{\scalebox{1.3}{$\frac{H}{8}\times \frac{W}{8}$}} & Patch Embedding & \multicolumn{4}{c}{$P_2=2$;\ \ \  $C_2=128$} \\
	\cline{3-7}
	& & \tabincell{c}{Transformer\\Encoder} &
	$\begin{bmatrix}
	\begin{array}{l}
	R_2=4 \\
	N_2=2 \\
	E_2=8 \\
	\end{array}
	\end{bmatrix} \times 2$ &
	$\begin{bmatrix}
	\begin{array}{l}
	R_2=4 \\
	N_2=2 \\
	E_2=8 \\
	\end{array}
	\end{bmatrix} \times 3$ &
	$\begin{bmatrix}
	\begin{array}{l}
	R_2=4 \\
	N_2=2 \\
	E_2=8 \\
	\end{array}
	\end{bmatrix} \times 3$ &
	$\begin{bmatrix}
	\begin{array}{l}
	R_2=4 \\
	N_2=2 \\
	E_2=8 \\
	\end{array}
	\end{bmatrix} \times 8$\\
	\hline
	\multirow{2}{*}[-2.5ex]{Stage 3} & \multirow{2}{*}[-2.5ex]{\scalebox{1.3}{$\frac{H}{16}\times \frac{W}{16}$}} & Patch Embedding & \multicolumn{4}{c}{$P_3=2$;\ \ \  $C_3=320$} \\
	\cline{3-7}
	& & \tabincell{c}{Transformer\\Encoder} &
	$\begin{bmatrix}
	\begin{array}{l}
	R_3=2 \\
	N_3=5 \\
	E_3=4 \\
	\end{array}
	\end{bmatrix} \times 2$ &
	$\begin{bmatrix}
	\begin{array}{l}
	R_3=2 \\
	N_3=5 \\
	E_3=4 \\
	\end{array}
	\end{bmatrix} \times 6$ &
	$\begin{bmatrix}
	\begin{array}{l}
	R_3=2 \\
	N_3=5 \\
	E_3=4 \\
	\end{array}
	\end{bmatrix} \times 18$ &
	$\begin{bmatrix}
	\begin{array}{l}
	R_3=2 \\
	N_3=5 \\
	E_3=4 \\
	\end{array}
	\end{bmatrix} \times 27$\\
	\hline
	\multirow{2}{*}[-2.5ex]{Stage 4} &  \multirow{2}{*}[-2.5ex]{\scalebox{1.3}{$\frac{H}{32}\times \frac{W}{32}$}} & Patch Embedding & \multicolumn{4}{c}{$P_4=2$;\ \ \ $C_4\!=\!512$} \\
	\cline{3-7}
	& & \tabincell{c}{Transformer\\Encoder} & 
	$\begin{bmatrix}
	\begin{array}{l}
	R_4=1 \\
	N_4=8 \\
	E_4=4 \\
	\end{array}
	\end{bmatrix} \times 2$ &
	$\begin{bmatrix}
	\begin{array}{l}
	R_4=1 \\
	N_4=8 \\
	E_4=4 \\
	\end{array}
	\end{bmatrix} \times 3$ & $\begin{bmatrix}
	\begin{array}{l}
	R_4=1 \\
	N_4=8 \\
	E_4=4 \\
	\end{array}
	\end{bmatrix} \times 3$ & $\begin{bmatrix}
	\begin{array}{l}
	R_4=1 \\
	N_4=8 \\
	E_4=4 \\
	\end{array}
	\end{bmatrix} \times 3$\\
\end{tabular}
    \caption{\textbf{Detailed settings of PVT series.} The design follows the two rules of ResNet~\cite{he2016deep}: (1) with the growth of network depth, the hidden dimension gradually increases, and the output resolution progressively shrinks; (2) the major computation resource is concentrated in Stage 3.
    }
    \label{tab:arch}
\end{table*}

\subsection{Discussion}

The most related work to our model is ViT~\cite{dosovitskiy2020image}. 
Here, we discuss the relationship and differences between them.
First, both PVT and ViT are pure Transformer models without convolutions. The primary difference between them is the pyramid structure. 
Similar to the traditional Transformer~\cite{vaswani2017attention}, the length of ViT's output sequence is the same as the input, which means that the output of ViT is single-scale (see Figure~\ref{fig:pipeline} (b)).
Moreover, due to the limited resource, the input of ViT is coarse-grained (\eg, the patch size is 16 or 32 pixels), and thus its output resolution is relatively low (\eg, 16-stride or 32-stride).
As a result, it is difficult to directly apply ViT to dense prediction tasks that require high-resolution or multi-scale feature maps.

Our PVT breaks the routine of Transformer by introducing a progressive shrinking pyramid.
It can generate multi-scale feature maps like a traditional CNN backbone.
In addition, we also designed a simple but
effective attention layer---SRA, to process high-resolution feature maps and reduce computational/memory costs.
Benefiting from the above designs, our method has the following advantages over ViT:
1) more flexible---can generate feature maps of different scales/channels in different stages;
2) more versatile---can be easily plugged and played in most downstream task models;
3) more friendly to computation/memory---can handle higher resolution feature maps or longer sequences.

\section{Application to Downstream Tasks}\label{sec:task}

\subsection{Image-Level Prediction}

Image classification is the most classical task of image-level prediction.
To provide instances for discussion, we design a series of PVT models with different scales, namely PVT-Tiny, -Small, -Medium, and -Large, whose parameter numbers are similar to ResNet18, 50, 101, and 152, respectively.
Detailed hyper-parameter settings of the PVT series are provided in the \emph{supplementary material (SM)}.

For image classification, we follow ViT~\cite{dosovitskiy2020image} and DeiT~\cite{touvron2020training} to append a learnable classification token to the input of the last stage, and then employ a fully connected (FC) layer to conduct classification on top of the token.

\subsection{Pixel-Level Dense Prediction}

In addition to image-level prediction, dense prediction that requires pixel-level classification or regression to be performed on the feature map, is also often seen in downstream tasks.
Here, we discuss two typical tasks, namely object detection, and semantic segmentation.

We apply our PVT models to three representative dense prediction methods, namely RetinaNet~\cite{lin2017focal}, Mask R-CNN~\cite{he2017mask}, and Semantic FPN~\cite{kirillov2019panoptic}.
RetinaNet is a widely used single-stage detector, Mask R-CNN is the most popular two-stage instance segmentation framework, and Semantic FPN is a vanilla semantic segmentation method without special operations (\eg, dilated convolution).
Using these methods as baselines enables us to adequately examine the effectiveness of different backbones.

The implementation details are as follows: 
(1) Like ResNet, we initialize the PVT backbone with the weights pre-trained on ImageNet;
(2) We use the output feature pyramid $\{F_1, F_2, F_3, F_4\}$ as the input of FPN~\cite{lin2017feature}, and then the refined feature maps are fed to the follow-up detection/segmentation head;
(3) When training the detection/segmentation model, none of the layers in PVT are frozen;
(4) Since the input for detection/segmentation can be an arbitrary shape, the position embeddings pre-trained on ImageNet may no longer be meaningful. Therefore, we perform bilinear interpolation on the pre-trained position embeddings according to the input resolution.

\section{Experiments}

We compare PVT with the two most representative CNN backbones, \ie, ResNet~\cite{he2016deep} and ResNeXt~\cite {xie2017aggregated}, which are widely used in the benchmarks of many downstream tasks.

\subsection{Image Classification}\label{sec:cls}

\noindent\textbf{Settings.}
Image classification experiments are performed on the ImageNet 2012 dataset \cite{russakovsky2015imagenet},
which comprises 1.28 million training images and 50K validation images from 1,000 categories. 
For fair comparison, all models are trained on the training set, and report the top-1 error on the validation set.
We follow DeiT~\cite{touvron2020training} and apply random cropping, random horizontal flipping \cite{szegedy2015going}, label-smoothing regularization \cite{szegedy2016rethinking}, mixup~\cite{zhang2017mixup}, CutMix~\cite{yun2019cutmix}, and random erasing~\cite{zhong2020random} as data augmentations.
During training, we employ AdamW~\cite{loshchilov2017decoupled} with a momentum of 0.9, a mini-batch size of 128, and a weight decay of $5\times 10^{-2}$ to optimize models. The initial learning rate is set to $1\times 10^{-3}$ and decreases following the cosine schedule~\cite{loshchilov2016sgdr}.  All models are trained for 300 epochs from scratch on 8 V100 GPUs.
To benchmark, we apply a center crop on the validation set, where a 224$\times$ 224 patch is cropped to evaluate the classification accuracy. 

\begin{table}[t]
    \centering
    \renewcommand\arraystretch{ 1.0}
    \setlength{\tabcolsep}{1.7mm}
    \footnotesize
    \begin{tabular}{l|c|c|c}
    \renewcommand{\arraystretch}{0.1}
	Method & \#Param (M) & GFLOPs & Top-1 Err (\%)  \\
	\whline
	ResNet18*~\cite{he2016deep} & 11.7 & 1.8 & 30.2  \\
	ResNet18~\cite{he2016deep}  & 11.7 & 1.8 & 31.5 \\
	DeiT-Tiny/16~\cite{touvron2020training} & 5.7 & 1.3 & 27.8 \\
	\rowcolor{mygray}
	PVT-Tiny (ours) & 13.2 & 1.9 &24.9 \\
	\hline
	ResNet50*~\cite{he2016deep}   &25.6 &4.1 &23.9 \\
	ResNet50~\cite{he2016deep}  &25.6 &4.1 & 21.5 \\
	ResNeXt50-32x4d*~\cite{xie2017aggregated} &25.0 &4.3 &22.4  \\
	ResNeXt50-32x4d~\cite{xie2017aggregated} &25.0 &4.3 & 20.5 \\
	T2T-ViT$_t$-14~\cite{t2tvit} & 22.0 & 6.1 & 19.3 \\
	TNT-S~\cite{tnt} & 23.8 & 5.2 & 18.7 \\
	DeiT-Small/16~\cite{touvron2020training}  & 22.1 & 4.6 & 20.1 \\
	\rowcolor{mygray}
	PVT-Small (ours)  & 24.5 & 3.8 & 20.2 \\
	\hline
	ResNet101*~\cite{he2016deep}  &44.7 & 7.9 &22.6\\
	ResNet101~\cite{he2016deep}  &44.7 & 7.9 & 20.2\\
	ResNeXt101-32x4d*~\cite{xie2017aggregated} & 44.2 & 8.0 &21.2 \\
	ResNeXt101-32x4d~\cite{xie2017aggregated} & 44.2  & 8.0 & 19.4 \\
	T2T-ViT$_t$-19~\cite{t2tvit} & 39.0 & 9.8 & 18.6 \\
	ViT-Small/16~\cite{dosovitskiy2020image} & 48.8 & 9.9 & 19.2 \\
	\rowcolor{mygray}
	PVT-Medium (ours) & 44.2 & 6.7 & 18.8\\
	\hline
	ResNeXt101-64x4d*~\cite{xie2017aggregated} & 83.5 & 15.6 & 20.4\\
	ResNeXt101-64x4d~\cite{xie2017aggregated} & 83.5 & 15.6 & 18.5\\
	ViT-Base/16~\cite{dosovitskiy2020image} & 86.6 & 17.6 & 18.2 \\
	T2T-ViT$_t$-24~\cite{t2tvit} & 64.0 & 15.0 & 17.8 \\
	TNT-B~\cite{tnt} & 66.0 & 14.1 & 17.2 \\
	DeiT-Base/16~\cite{touvron2020training} & 86.6 & 17.6 & 18.2 \\
	\rowcolor{mygray}
	PVT-Large (ours) & 61.4 & 9.8 & 18.3 \\
\end{tabular}
    \caption{\textbf{Image classification performance on the ImageNet validation set}.
    ``\#Param'' refers to the number of parameters. 
    ``GFLOPs'' is calculated under the input scale of $224\times 224$. ``*'' indicates the performance of the method trained under the strategy of its original paper.}
    \label{tab:cls}
\end{table}
\noindent\textbf{Results.}
In Table~\ref{tab:cls}, we see that our PVT models are superior to conventional CNN backbones under similar parameter numbers and computational budgets.
For example, when the GFLOPs are roughly similar, the top-1 error of PVT-Small reaches 20.2, which is 1.3 points higher than that of ResNet50~\cite{he2016deep} (20.2 \vs 21.5).
Meanwhile, under similar or lower complexity, PVT models archive performances comparable to the recently proposed Transformer-based models, such as ViT~\cite{dosovitskiy2020image} and DeiT~\cite{touvron2020training} (PVT-Large: 18.3 \vs ViT(DeiT)-Base/16: 18.3).
Here, we clarify that these results are within our expectations, because the pyramid structure is beneficial to dense prediction tasks, but
brings little improvements to image classification.

Note that ViT and DeiT have limitations as they are specifically designed for classification tasks, and thus are not suitable for dense prediction tasks, which usually require effective feature pyramids.

\subsection{Object Detection}\label{sec:det}

\noindent\textbf{Settings.}
Object detection experiments are conducted on the challenging COCO benchmark~\cite{lin2014microsoft}.
All models are trained on COCO \texttt{train2017} (118k images) and evaluated on \texttt{val2017} (5k images).
We verify the effectiveness of PVT backbones on top of two standard detectors, namely RetinaNet~\cite{lin2017focal} and Mask R-CNN~\cite{he2017mask}.
Before training, we use the weights pre-trained on ImageNet to initialize the backbone and Xavier~\cite{glorot2010understanding} to initialize the newly added layers.
Our models are trained with a batch size of 16 on 8 V100 GPUs and optimized by AdamW~\cite{loshchilov2017decoupled} with an initial learning rate of $1\times10^{-4}$.
Following common practices~ \cite{lin2017focal,he2017mask,chen2019mmdetection}, we adopt 1$\times$ or 3$\times$ training schedule~(\ie, 12 or 36 epochs) to train all detection models.
The training image is resized to have a shorter side of 800 pixels, while the longer side does not exceed 1,333 pixels.
When using the 3$\times$ training schedule, we randomly resize the shorter side of the input image within the range of $[640, 800]$.
In the testing phase, the shorter side of the input image is fixed to 800 pixels.

\begin{table*}[t]
    \centering
    \renewcommand\arraystretch{ 1.0}
    \setlength{\tabcolsep}{1.9mm}
    \footnotesize

\begin{tabular}{l|c|lcc|ccc|lcc|ccc}
\renewcommand{\arraystretch}{0.1}
\multirow{2}{*}{Backbone} & \multirow{2}{*}{\tabincell{c}{\#Param \\(M)}} &\multicolumn{6}{c|}{RetinaNet 1x} &\multicolumn{6}{c}{RetinaNet 3x + MS} \\
\cline{3-14} 
& &AP &AP$_{50}$ &AP$_{75}$ &AP$_S$ &AP$_M$ &AP$_L$ &AP &AP$_{50}$ &AP$_{75}$ &AP$_S$ &AP$_M$ &AP$_L$ \\
\whline
ResNet18~\cite{he2016deep} &{21.3} & 31.8 & 49.6 & 33.6 & 16.3 & 34.3 & 43.2 & 35.4 & 53.9 & 37.6 & 19.5 & 38.2  &46.8 \\
\rowcolor{mygray}
PVT-Tiny (ours) &23.0& {36.7}\gbf{+4.9}& {56.9}& {38.9}& {22.6}& {38.8} &{50.0} & {39.4}\gbf{+4.0} & {59.8} & {42.0} & {25.5} & {42.0} & {52.1} \\
\hline
ResNet50~\cite{he2016deep} &37.7 & 36.3 & 55.3 & 38.6 & 19.3 & 40.0 & 48.8 & 39.0 & 58.4 & 41.8 & 22.4 & 42.8 & 51.6  \\
\rowcolor{mygray}
PVT-Small (ours) & {34.2} & {40.4}\gbf{+4.1} & {61.3} & {43.0} & {25.0} & {42.9} & {55.7} & {42.2}\gbf{+3.2} & {62.7} & {45.0} & {26.2} & {45.2} & {57.2 }\\
\hline
ResNet101~\cite{he2016deep} &56.7  & 38.5 & 57.8 & 41.2 & 21.4 & 42.6 & 51.1 & 40.9 & 60.1 & 44.0 & 23.7 & 45.0 & 53.8
\\
ResNeXt101-32x4d~\cite{xie2017aggregated} &56.4& 39.9\gbf{+1.4} & 59.6 & 42.7 & 22.3 & 44.2 & 52.5 & 41.4\gbf{+0.5} & 61.0 & 44.3 &23.9& 45.5 & 53.7 \\
\rowcolor{mygray}
PVT-Medium (ours) &{53.9} & {41.9}\gbf{+3.4} & {63.1} & {44.3} & {25.0} & {44.9} & {57.6} & {43.2}\gbf{+2.3} & {63.8} & {46.1} & {27.3} & {46.3} & {58.9}\\
\hline
ResNeXt101-64x4d~\cite{xie2017aggregated} & 95.5& 41.0 & 60.9 & 44.0 & 23.9 & 45.2 & 54.0 & 41.8& 61.5 & 44.4 & 25.2& 45.4& 54.6 \\
\rowcolor{mygray}
PVT-Large (ours) & 71.1 & {42.6}\gbf{+1.6} & {63.7} & {45.4} & {25.8} & {46.0} & {58.4} & 43.4\gbf{+1.6} &63.6& 46.1& 26.1& 46.0& 59.5
 \\
\end{tabular}
     
    \caption{\textbf{Object detection performance on COCO \texttt{val2017}.} 
    ``MS'' means that multi-scale training~\cite{lin2017focal,he2017mask} is used.}
    \label{tab:retina} 
\end{table*}

\begin{table*}[t]
    \centering
    \renewcommand\arraystretch{.9}
    \setlength{\tabcolsep}{1.4mm}
    \footnotesize

\begin{tabular}{l|c|lcc|lcc|lcc|lcc}
\renewcommand{\arraystretch}{0.1}
\multirow{2}{*}{Backbone} & \multirow{2}{*}{\tabincell{c}{\#Param \\(M)}} &\multicolumn{6}{c|}{Mask R-CNN 1x} &\multicolumn{6}{c}{Mask R-CNN 3x + MS} \\
\cline{3-14} 
& &AP$^{\rm b}$ &AP$_{50}^{\rm b}$ &AP$_{75}^{\rm b}$ &AP$^{\rm m}$ &AP$_{50}^{\rm m}$ &AP$_{75}^{\rm m}$ &AP$^{\rm b}$ &AP$_{50}^{\rm b}$ &AP$_{75}^{\rm b}$ &AP$^{\rm m}$ &AP$_{50}^{\rm m}$ &AP$_{75}^{\rm m}$ \\
\whline
ResNet18~\cite{he2016deep} & {31.2} & 34.0 & 54.0 & 36.7 & 31.2 & 51.0 & 32.7 & 36.9 & 57.1 & 40.0 & 33.6 & 53.9 & 35.7\\
\rowcolor{mygray}
PVT-Tiny (ours) & 32.9 & {36.7}\gbf{+2.7} & {59.2} & {39.3} & {35.1}\gbf{+3.9} & {56.7} & {37.3} & {39.8}\gbf{+2.9} & {62.2} & {43.0} & {37.4}\gbf{+3.8} & {59.3} & {39.9} \\
\hline
ResNet50~\cite{he2016deep} & 44.2& 38.0 & 58.6 & 41.4 & 34.4 & 55.1 & 36.7 & 41.0 & 61.7 & 44.9 & 37.1 & 58.4 & 40.1\\
\rowcolor{mygray}
PVT-Small (ours) &{44.1} & {40.4}\gbf{+2.4} & {62.9} & {43.8} & {37.8}\gbf{+3.4} & {60.1} & {40.3} & {43.0}\gbf{+2.0} & {65.3} & {46.9} & {39.9}\gbf{+2.8} & {62.5} & {42.8}\\
\hline
ResNet101~\cite{he2016deep} & 63.2 & 40.4 & 61.1 & 44.2 & 36.4 & 57.7 & 38.8 & 42.8 & 63.2 & 47.1 & 38.5& 60.1& 41.3\\
ResNeXt101-32x4d~\cite{xie2017aggregated} &{62.8} & 41.9\gbf{+1.5} & 62.5 & {45.9} & 37.5\gbf{+1.1} & 59.4 & 40.2 & 44.0\gbf{+1.2} & 64.4 & 48.0 & 39.2\gbf{+0.7} & 61.4 & 41.9 \\
\rowcolor{mygray}
PVT-Medium (ours) &63.9 & {42.0}\gbf{+1.6} &{64.4} &45.6 &{39.0}\gbf{+2.6}& {61.6}& {42.1} & {44.2}\gbf{+1.4} & {66.0} & {48.2} & {40.5}\gbf{+2.0} & {63.1} & {43.5}\\
\hline
ResNeXt101-64x4d~\cite{xie2017aggregated} &101.9 & 42.8 & 63.8 & {47.3} & 38.4 & 60.6 & 41.3 & 44.4 & 64.9 & 48.8 & 39.7& 61.9 & 42.6 \\
\rowcolor{mygray}
PVT-Large (ours) &81.0 & {42.9}\gbf{+0.1}& {65.0} & 46.6 &{39.5}\gbf{+1.1}& {61.9}& {42.5} & 44.5\gbf{+0.1} &66.0& 48.3 &40.7\gbf{+1.0} &63.4& 43.7 \\
\end{tabular}
     
    \caption{\textbf{Object detection and instance segmentation performance on COCO \texttt{val2017}.}
    AP$^{\rm b}$ and AP$^{\rm m}$ denote bounding box AP and mask AP, respectively. 
    }
    \label{tab:mask}.
\end{table*}

\noindent\textbf{Results.}
As shown in Table~\ref{tab:retina}, when using RetinaNet for object detection, we find that under comparable number of parameters, the PVT-based models significantly surpasses their counterparts. 
For example, with the 1$\times$ training schedule, the AP of PVT-Tiny is 4.9 points better than that of ResNet18~(36.7 \vs 31.8).
Moreover, with the 3$\times$ training schedule and multi-scale training, PVT-Large archive the best AP of 43.4, surpassing ResNeXt101-64x4d (43.4 \vs 41.8), while our parameter number is 30\% fewer.
These results indicate that our PVT can be a good alternative to the CNN backbone for object detection.

Similar results are found in instance segmentation experiments based on Mask R-CNN, as shown in Table~\ref{tab:mask}.
With the 1$\times$ training schedule, PVT-Tiny achieves 35.1 mask AP (AP$^{\rm m}$), which is 3.9 points better than ResNet18 (35.1 \vs 31.2) and even 0.7 points higher than ResNet50 (35.1 \vs 34.4). 
The best AP$^{\rm m}$ obtained by PVT-Large is 40.7, which is 1.0 points higher than ResNeXt101-64x4d (40.7 \vs 39.7), with 20\% fewer parameters.

\subsection{Semantic Segmentation}\label{sec:seg}

\begin{table}[t]
    \centering
    \renewcommand\arraystretch{ 1.0}
    \setlength{\tabcolsep}{2.4mm}
    \footnotesize
    \begin{tabular}{l|c|c|l}
\renewcommand{\arraystretch}{0.1}
	\multirow{2}{*}{Backbone} & \multicolumn{3}{c}{Semantic FPN}\\
	\cline{2-4}
	& \#Param (M) & GFLOPs & mIoU (\%)   \\
	\whline
	ResNet18~\cite{he2016deep} & {15.5} & 32.2 & 32.9 \\
	\rowcolor{mygray}
	PVT-Tiny (ours) & 17.0 & 33.2 & {35.7}\gbf{+2.8} \\
	\hline
ResNet50~\cite{he2016deep} & 28.5&45.6 & 36.7\\
\rowcolor{mygray}
PVT-Small (ours) & {28.2}& 44.5& {39.8}\gbf{+3.1}\\
\hline
ResNet101~\cite{he2016deep} & 47.5&65.1& 38.8\\
ResNeXt101-32x4d~\cite{xie2017aggregated} & {47.1} &64.7& 39.7\gbf{+0.9} \\
\rowcolor{mygray}
PVT-Medium (ours) & 48.0 &61.0& {41.6}\gbf{+2.8}\\
\hline
ResNeXt101-64x4d~\cite{xie2017aggregated} & 86.4 &103.9& 40.2\\
\rowcolor{mygray}
PVT-Large (ours) & {65.1} &79.6& {42.1}\gbf{+1.9} \\
\hline
\rowcolor{mygray}
PVT-Large{*} (ours) & 65.1 &79.6& {44.8} \\
\end{tabular}
     
    \caption{\textbf{Semantic segmentation performance of different backbones on the ADE20K validation set.}
    ``GFLOPs'' is calculated under the input scale of $512\times 512$.
    ``*'' indicates 320K iterations training and multi-scale flip testing.
    }
    \label{tab:seg}
\end{table}

\noindent\textbf{Settings.}
We choose ADE20K~\cite{zhou2017scene}, a challenging scene parsing dataset, to benchmark the performance of semantic segmentation. ADE20K contains 150 fine-grained semantic categories, with 20,210, 2,000, and 3,352 images for training, validation, and testing, respectively. 
We evaluate our PVT backbones on the basis of Semantic FPN~\cite{kirillov2019panoptic}, a simple segmentation method without dilated convolutions~\cite{yu2015multi}.
In the training phase, the backbone is initialized with the weights pre-trained on ImageNet~\cite{deng2009imagenet}, and other newly added layers are initialized with Xavier~\cite{glorot2010understanding}. 
We optimize our models using AdamW~\cite{loshchilov2017decoupled} with an initial learning rate of 1e-4.
Following common practices~\cite{kirillov2019panoptic,chen2017deeplab}, we train our models for 80k iterations with a batch size of 16 on 4 V100 GPUs.
The learning rate is decayed following the polynomial decay schedule with a power of 0.9.
We randomly resize and crop the image to $512\times 512$ for training, and rescale to have a shorter side of 512 pixels during testing.

\noindent\textbf{Results.} 
As shown in Table~\ref{tab:seg}, when using Semantic FPN~\cite{kirillov2019panoptic} for semantic segmentation, PVT-based models consistently outperforms the models based on ResNet~\cite{he2016deep} or ResNeXt~\cite{xie2017aggregated}.
For example, with almost the same number of parameters and GFLOPs, our PVT-Tiny/Small/Medium are at least 2.8 points higher than ResNet-18/50/101. 
In addition, although the parameter number and GFLOPs of our PVT-Large are 20\% lower than those of ResNeXt101-64x4d,
the mIoU is still 1.9 points higher~(42.1 \vs 40.2).
With a longer training schedule and multi-scale testing, PVT-Large+Semantic FPN archives the best mIoU of 44.8, which is very close to the state-of-the-art performance of the ADE20K benchmark. Note that Semantic FPN is just a simple segmentation head.
These results demonstrate that our PVT backbones can extract better features for semantic segmentation than the CNN backbone, benefiting from the global attention mechanism.

\subsection{Pure Transformer Detection \& Segmentation}

\begin{table}[t]
    \centering
    \renewcommand\arraystretch{ 1.0}
    \setlength{\tabcolsep}{1.4mm}
    \footnotesize
    \begin{tabular}{l|lcc|ccc}
    \renewcommand{\arraystretch}{0.1}
	\multirow{2}{*}{Method} & \multicolumn{6}{c}{DETR (50 Epochs)}\\
	\cline{2-7}
 &AP &AP$_{50}$ &AP$_{75}$ &AP$_S$ &AP$_M$ &AP$_L$  \\
	\whline
	ResNet50~\cite{he2016deep} & 32.3 & 53.9 & 32.3 & 10.7 & 33.8 & 53.0  \\
	\rowcolor{mygray}
	PVT-Small (ours) & 34.7\gbf{+2.4} & 55.7 & 35.4 & 12.0 & 36.4 & 56.7 \\
\end{tabular}
     
    \caption{\textbf{Performance of the pure Transformer object detection pipeline.} We build a pure Transformer detector by combining PVT and DETR~\cite{carion2020end}, whose AP is 2.4 points higher than the original DETR based on ResNet50~\cite{he2016deep}.}
    \label{tab:detr}
\end{table}
\noindent\textbf{PVT+DETR.} To reach the limit of no convolution, we build a pure Transformer pipeline for object detection by simply combining our PVT with a Transformer-based detection head---DETR~\cite{carion2020end}. We train models on COCO \texttt{train2017} for 50 epochs with an initial learning rate of $1\times 10^{-4}$. The learning rate is divided by 10 at the 33rd epoch. We use random flipping and multi-scale training as data augmentation. All other experimental settings is the same as those in Sec.~\ref{sec:det}. As reported in Table \ref{tab:detr}, PVT-based DETR archieves 34.7 AP on COCO \texttt{val2017}, outperforming the original ResNet50-based DETR by 2.4 points (34.7 \vs 32.3). These results prove that \emph{a pure Transformer detector can also works well 
in the object detection task}.

\begin{table}[t]
    \centering
    \footnotesize
    \renewcommand\arraystretch{ 1.0}
    \setlength{\tabcolsep}{1.3mm}
    \begin{tabular}{l|c|c|l}
    \renewcommand{\arraystretch}{0.1}
	Method & \#Param (M) &GFLOPs & mIoU (\%)   \\
	\whline
	ResNet50-d8+DeeplabV3+~\cite{chen2018encoder}  &26.8&120.5 &41.5\\
	ResNet50-d16+DeeplabV3+~\cite{chen2018encoder}  &26.8&45.5 &40.6\\
	\hline
	ResNet50-d16+Trans2Seg~\cite{xie2021segmenting}  & 56.1&79.3 & 39.7\\
	\rowcolor{mygray} PVT-Small+Trans2Seg   &32.1 &31.6&42.6\gbf{+2.9}\\
	
\end{tabular}
     
    \caption{\textbf{Performance of the pure Transformer semantic segmentation pipeline.} We build a pure Transformer detector by combining PVT and Trans2Seg~\cite{xie2021segmenting}. It is 2.9\% higher than ResNet50-d16+Trans2Seg and 1.1\% higher than ResNet50-d8+DeeplabV3+ with lower GFlops. ``d8'' and ``d16'' means dilation 8 and 16, respectively. }
    \label{tab:trans2seg}
\end{table}
\noindent\textbf{PVT+Trans2Seg.}We build a pure Transformer model for semantic segmentation by combining our PVT with Trans2Seg~\cite{xie2021segmenting}, a Transformer-based segmentation head.
According to the experimental settings in Sec.~5.3, we perform experiments on ADE20K~\cite{zhou2017scene} with 40k iterations training, single scale testing, and compare it with ResNet50+Trans2Seg~\cite{xie2021segmenting} and DeeplabV3+~\cite{chen2018encoder} with ResNet50-d8 (dilation 8) and -d16(dilation 8) in Table \ref{tab:trans2seg}. 
We find that our PVT-Small+Trans2Seg achieves 42.6 mIoU, outperforming ResNet50-d8+DeeplabV3+~(41.5).
Note that, ResNet50-d8+DeeplabV3+ has 120.5 GFLOPs due to the high computation cost of dilated convolution, and our method has only 31.6 GFLOPs, which is 4 times fewer.
In addition, our PVT-Small+Trans2Seg performs better than ResNet50-d16+Trans2Seg~(mIoU: 42.6 \vs 39.7, GFlops: 31.6 \vs 79.3).
These results prove that \emph{a pure Transformer segmentation network is workable.}

\subsection{Ablation Study}

\noindent\textbf{Settings.}
We conduct ablation studies on ImageNet~\cite{deng2009imagenet} and COCO~\cite{lin2014microsoft} datasets.
The experimental settings on ImageNet are the same as the settings in Sec. \ref{sec:cls}.
For COCO, all models are trained with a 1$\times$ training schedule (\ie, 12 epochs) and without multi-scale training, and other settings follow those in Sec. \ref{sec:det}.

\begin{table}[t]
    \centering
    \renewcommand\arraystretch{ 1.0}
    \setlength{\tabcolsep}{1.1mm}
    \footnotesize
    \begin{tabular}{l|c|ccc|ccc}
    \renewcommand{\arraystretch}{0.1}
	\multirow{2}{*}{Method} & \multirow{2}{*}{\tabincell{c}{\#Param \\(M)}} & \multicolumn{6}{c}{RetinaNet 1x}\\
	\cline{3-8}
	&&AP &AP$_{50}$ &AP$_{75}$ &AP$_S$ &AP$_M$ &AP$_L$  \\
	\whline
	ViT-Small/4~\cite{dosovitskiy2020image} & 60.9 &\multicolumn{6}{c}{Out of Memory} \\
	ViT-Small/32~\cite{dosovitskiy2020image} & 60.8 & 31.7 & 51.3 & 32.3 &14.8&33.7 &47.9 \\
	PVT-Small (ours) & {34.2} & {40.4} & {61.3} & {43.0} & 25.0 & 42.9 & 55.7 \\
\end{tabular}
     
    \caption{\textbf{Performance comparison between ViT and our PVT using  RetinaNet for object detection.} ViT-Small/4 runs out of GPU memory due to small patch size (\ie, $4\!\times\!4$ per patch). ViT-Small/32 obtains 31.7 AP on COCO \texttt{val2017},  which is 8.7 points lower than our PVT-Small.}
    \label{tab:vit_pvt_det}
\end{table}

\noindent\textbf{Pyramid Structure.} A Pyramid structure is crucial when applying Transformer to dense prediction tasks. ViT (see Figure \ref{fig:pipeline} (b)) is a columnar framework, whose output is single-scale. This results in a low-resolution output feature map when using coarse image patches (\eg, 32$\times$32 pixels per patch) as input,
leading to poor detection performance (31.7 AP on COCO \texttt{val2017}),\footnote{For adapting ViT to RetinaNet, we extract the features from the layer 2, 4, 6, and 8 of ViT-Small/32, and interpolate them to different scales.} as shown in Table \ref{tab:vit_pvt_det}. When using fine-grained image patches (\eg, 4$\times$4 pixels per patch) as input like our PVT, ViT will exhaust the GPU memory (32G).
Our method avoids this problem through a progressive shrinking pyramid. Specifically, our model can process high-resolution feature maps in shallow stages and low-resolution feature maps in deep stages.
Thus, it obtains a promising AP of 40.4 on COCO \texttt{val2017}, 8.7 points higher than ViT-Small/32 (40.4 \vs 31.7).

\begin{figure}
		\centering
		\setlength{\fboxrule}{0pt}
		\fbox{\includegraphics[width=0.45\textwidth]{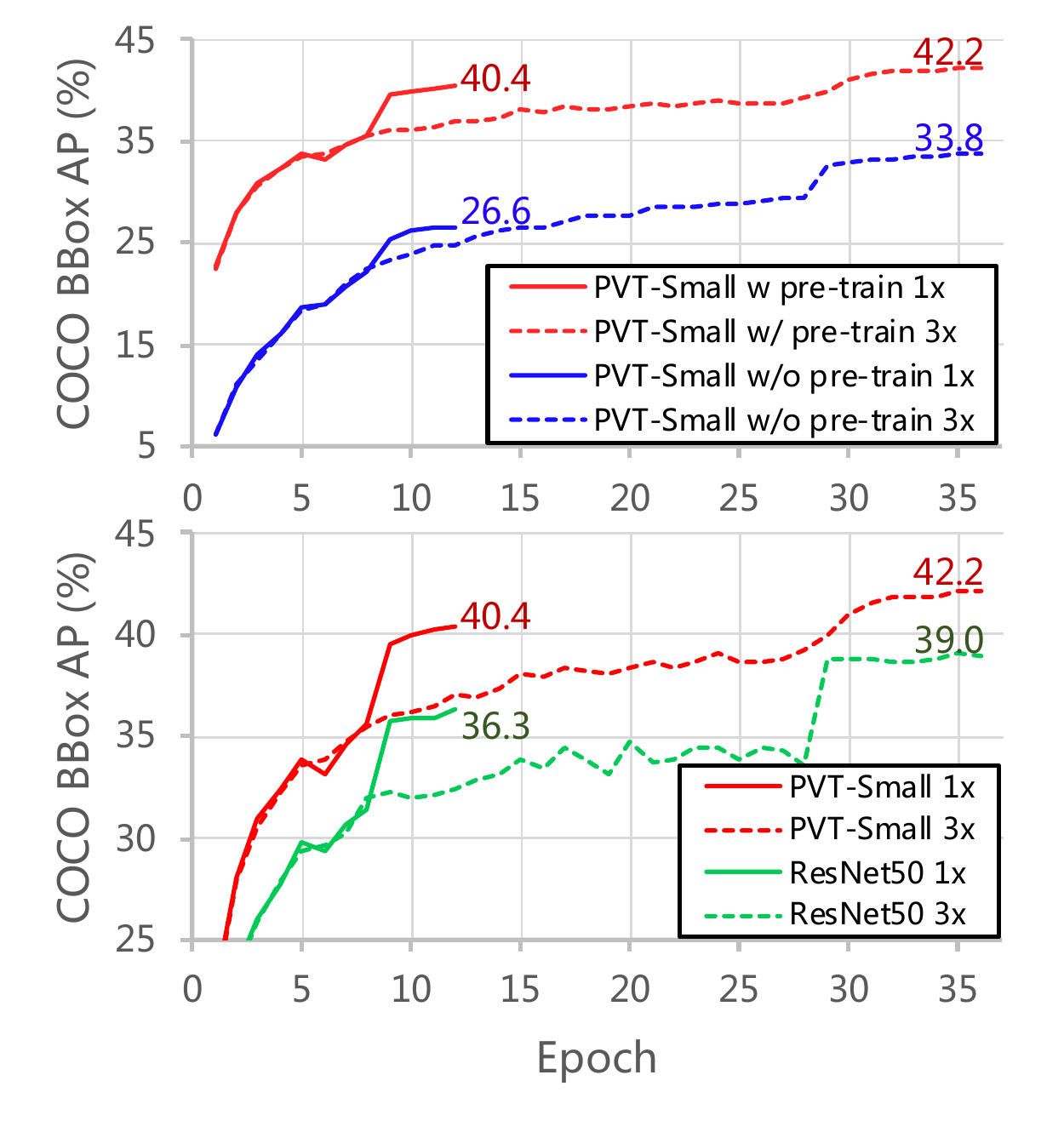}}
		\caption{\textbf{AP curves of RetinaNet on COCO \texttt{val2017} under different backbone settings.} Top: using weights pre-trained on ImageNet \vs random initialization. Bottom: PVT-S \vs R50~\cite{he2016deep}.}
		\label{fig:pretrain}
\end{figure}

\begin{table}[t]
    \centering
    \renewcommand\arraystretch{ 1.0}
    \footnotesize
    \begin{tabular}{l|c|c|ccc}
	\multirow{2}{*}{Method} & \multirow{2}{*}{\tabincell{c}{\#Param \\(M)}} & \multirow{2}{*}{Top-1} & \multicolumn{3}{c}{RetinaNet 1x}\\
	\cline{4-6}
	& & & AP &AP$_{50}$ &AP$_{75}$  \\
	\whline
	Wider PVT-Small & 46.8 & 19.3 & 40.8 & 61.8 &43.3 \\
	Deeper PVT-Small & 44.2 & {18.8} & {41.9} & {63.1} & {44.3}  \\
\end{tabular}
    \caption{\textbf{Deeper \vs Wider.} ``Top-1'' denotes the top-1 error on the ImageNet validation set. ``AP'' denotes the bounding box AP on COCO \texttt{val2017}. The deep model (\ie, PVT-Medium) obtains better performance than the wide model (\ie, PVT-Small-Wide ) under comparable parameter number.}
    \label{tab:deep_wide}
\end{table}

\noindent\textbf{Deeper \vs Wider.} 
The problem of whether the CNN backbone should go deeper or wider has been extensively discussed in previous work~\cite{he2016deep,zerhouni2017wide}. Here, we explore this problem in our PVT.
For fair comparisons, we multiply the hidden dimensions $\{C_1, C_2, C_3, C_4\}$ of PVT-Small
by a scale factor 1.4 to make it have an equivalent parameter number to the deep model (\ie, PVT-Medium). As shown in Table \ref{tab:deep_wide}, the deep model (\ie, PVT-Medium) consistently works better than the wide model (\ie, PVT-Small-Wide) on both ImageNet and COCO. Therefore, going deeper is more effective than going wider in the design of PVT. Based on this observation, in Table~\ref{tab:arch}, we develop PVT models with different scales by increasing the model depth.

\noindent\textbf{Pre-trained Weights.} Most dense prediction models (\eg, RetinaNet~\cite{lin2017focal}) rely on the backbone whose weights are pre-trained on ImageNet. We also discuss this problem in our PVT. In the top of Figure \ref{fig:pretrain}, we plot the validation AP curves of RetinaNet-PVT-Small w/ ({red curves}) and w/o ({blue curves}) pre-trained weights.
We find that the model w/ pre-trained weights converges better than the one w/o pre-trained weights, 
and the gap between their final AP reaches 13.8 under the 1$\times$ training schedule and 8.4 under the 3$\times$ training schedule and multi-scale training.
Therefore, like CNN-based models, pre-training weights can also help PVT-based models converge faster and better.
Moreover, in the bottom of Figure \ref{fig:pretrain}, we also see that the convergence speed of PVT-based models ({red curves}) is faster than that of ResNet-based models ({green curves}).

\begin{table}[t]
    \centering
    \footnotesize
    \setlength{\tabcolsep}{1.8mm}
    \begin{tabular}{l|c|c|ccc}
	\multirow{2}{*}{Method} & \multirow{2}{*}{\tabincell{c}{\#Param \\(M)}} & \multirow{2}{*}{GFLOPs} & \multicolumn{3}{c}{Mask R-CNN 1x}\\
	\cline{4-6}
	& & & AP$^{\rm m}$ &AP$_{50}^{\rm m}$ &AP$_{75}^{\rm m}$  \\
	\whline
	ResNet50+GC r4~\cite{cao2019gcnet} &54.2 &279.6 & 36.2 & 58.7 & 38.3  \\
	PVT-Small (ours) & 44.1 & 304.4 & 37.8 & 60.1 &40.3 \\
\end{tabular}
    \caption{\textbf{PVT \vs CNN w/ non-local.} AP$^{\rm m}$ denotes mask AP. Under similar parameter nubmer and GFLOPs, our PVT outperform the CNN backbone w/ Non-Local (ResNet50+GC r4) by 1.6 AP$^{\rm m}$ (37.8 \vs 36.2).}
    \label{tab:non_local}
\end{table}

\noindent\textbf{PVT \vs ``CNN w/ Non-Local''}
To obtain a global receptive field, some well-engineered CNN backbones, such as GCNet~\cite{cao2019gcnet}, integrate the non-local block in the CNN framework.
Here, we compare the performance of our PVT (pure Transformer) and GCNet (CNN w/ non-local), using Mask R-CNN for instance segmentation.
As reported in Table \ref{tab:non_local},  we find that our PVT-Small outperforms ResNet50+GC r4~\cite{cao2019gcnet} by 1.6 points in AP$^{\rm m}$ (37.8 \vs 36.2), and 2.0 points in AP$_{75}^{\rm m}$ (38.3 \vs 40.3), under comparable parameter number and GFLOPs. 
There are two possible reasons for this result:

(1) Although a single global attention layer (\eg, non-local~\cite{wang2018non} or multi-head attention (MHA)~\cite{vaswani2017attention}) can acquire global-receptive-field features, the model performance keeps improving as the model deepens.
This indicates that \emph{stacking multiple MHAs can further enhance the representation capabilities of features.}
Therefore, as a pure Transformer backbone with more global attention layers, our PVT tends to perform better than the CNN backbone equipped with non-local blocks (\eg, GCNet).

(2) Regular convolutions can be deemed
as special instantiations of spatial attention mechanisms~\cite{zhu2019empirical}. In other words, the format of MHA is more flexible than the regular convolution.
For example, for different inputs, the weights of the convolution are fixed, but the attention weights of MHA change dynamically with the input.
Thus, \emph{the features learned by the pure Transformer backbone full of MHA layers, could be more flexible and expressive.}

\begin{figure}
		\centering
		\setlength{\fboxrule}{0pt}
		\fbox{\includegraphics[width=0.45\textwidth]{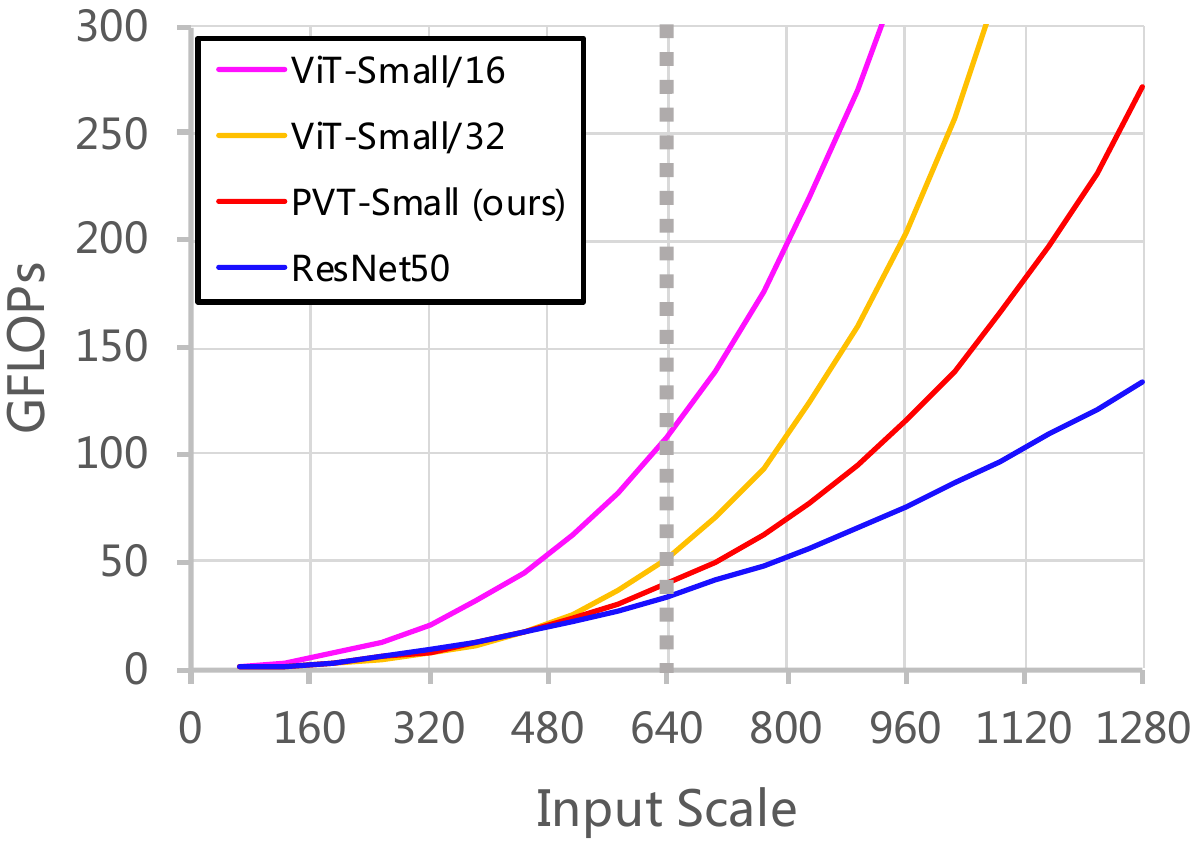}}
		 
		\caption{\textbf{Models' GFLOPs under different input scales.} The growth rate of GFLOPs: ViT-Small/16~\cite{dosovitskiy2020image}$>$ViT-Small/32~\cite{dosovitskiy2020image}$>$PVT-Small (ours)$>$ResNet50~\cite{he2016deep}.
		When the input scale is less than $640\times 640$, the GFLOPs of PVT-Small and ResNet50~\cite{he2016deep} are similar.
		}
		\label{fig:flops}
\end{figure}

\noindent\textbf{Computation Overhead.}
\begin{table}[t]
    \centering
    \renewcommand\arraystretch{ 1.0}
    \setlength{\tabcolsep}{1.6mm}
    \footnotesize
    \begin{tabular}{l|c|c|c|ccc}
    \renewcommand{\arraystretch}{0.1}
	\multirow{2}{*}{Method} &  
	\multirow{2}{*}{Scale} & 
	\multirow{2}{*}{GFLOPs} & \multirow{2}{*}{\tabincell{c}{Time\\(ms)}} & \multicolumn{3}{c}{RetinaNet 1x}\\
	\cline{5-7}
	&&&&AP &AP$_{50}$ &AP$_{75}$  \\
	\whline
	ResNet50~\cite{he2016deep} &800&239.3&55.9 & 36.3 & 55.3 & 38.6\\
	\hline
	\multirow{2}{*}{PVT-Small (ours)} & 640 &157.2& 51.7  & 38.7 & 59.3 & 40.8 \\
	 & 800 &285.8 &76.9 & 40.4 & 61.3 & {43.0} \\
	
\end{tabular}
     
    \caption{\textbf{Latency and AP under different input scales.} ``Scale'' and ``Time'' denote the input scale and time cost per image. When the shorter side 
    is 640 pixels, the PVT-Small+RetinaNet has a lower 
    GFLOPs and time cost (on a V100 GPU) than ResNet50+RetinaNet, while obtaining 2.4 points better AP (38.7 \vs 36.3).
    }
    \label{tab:speed}
\end{table}
With increasing input scale, the growth rate of the GFLOPs of our PVT is greater than ResNet~\cite{he2016deep}, but lower than ViT~\cite{dosovitskiy2020image}, as shown in Figure \ref{fig:flops}.
However, when the input scale does not exceed 640$\times$640 pixels, the GFLOPs of PVT-Small and ResNet50 are similar.
This means that our PVT is more suitable for tasks with medium-resolution input.

On COCO, the shorter side of the input image is 800 pixels. Under this condition, the inference speed of RetinaNet based on PVT-Small is slower than the ResNet50-based model, as reported in Table \ref{tab:speed}.
(1) \emph{A direct solution for this problem is to reduce the input scale.} When reducing the shorter side of the input image to 640 pixels, the model based on PVT-Small runs faster than the ResNet50-based model (51.7ms \vs, 55.9ms), with 2.4 higher AP (38.7 \vs 36.3).
2) \emph{Another solution is to develop a self-attention layer with lower computational complexity.} This is a worth exploring direction, we recently propose a solution PVTv2~\cite{wang2021pvtv2}.

\begin{figure*}[t]
		\centering
		\setlength{\fboxrule}{0pt}
		\fbox{\includegraphics[width=0.98\textwidth]{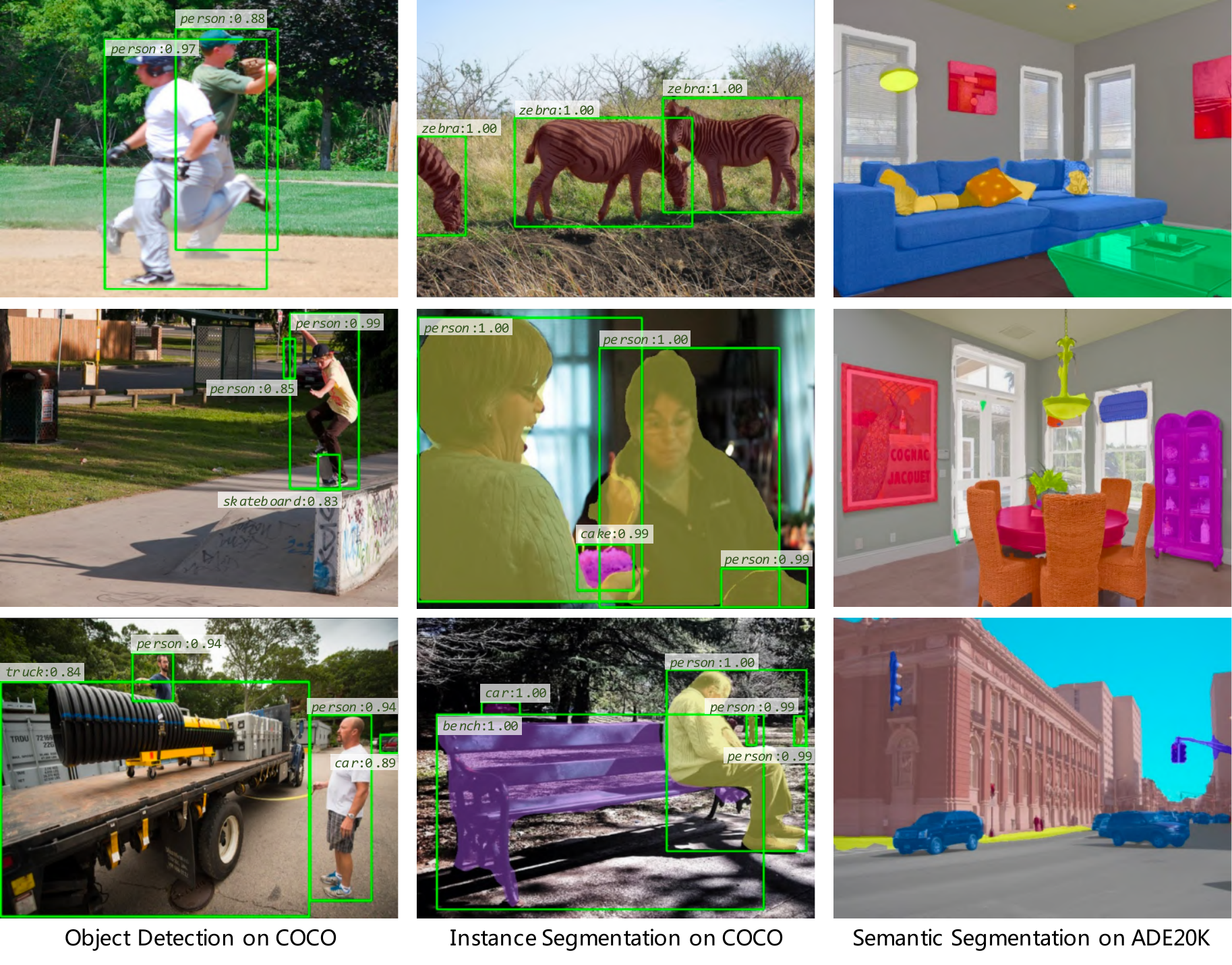}}
		\caption{\textbf{Qualitative results of object detection and instance segmentation on COCO \texttt{val2017}~\cite{lin2014microsoft}, and semantic segmentation on ADE20K~\cite{zhou2017scene}.} The results (from left to right) are generated by PVT-Small-based RetinaNet~\cite{lin2017focal}, Mask R-CNN~\cite{he2017mask}, and Semantic FPN~\cite{kirillov2019panoptic}, respectively.}
		\label{fig:res}
\end{figure*}

\noindent\textbf{Detection \& Segmentation Results.}
In Figure \ref{fig:res}, we also present some qualitative object detection and instance segmentation results on COCO \texttt{val2017}~\cite{lin2014microsoft}, and semantic segmentation results on ADE20K~\cite{zhou2017scene}. These results indicate that a pure Transformer backbone (\ie, PVT) without convolutions can also be easily plugged in dense prediction models (\eg, RetinaNet~\cite{lin2017focal}, Mask R-CNN~\cite{he2017mask}, and Semantic FPN~\cite{kirillov2019panoptic}), and obtain high-quality results.

\section{Conclusions and Future Work}

We introduce PVT, a pure Transformer backbone for dense prediction tasks, such as object detection and semantic segmentation. 
We develop a progressive shrinking pyramid and a spatial-reduction attention layer to obtain high-resolution and multi-scale feature maps under limited computation/memory resources.
Extensive experiments on object detection and semantic segmentation benchmarks verify that our PVT is stronger than well-designed CNN backbones under comparable numbers of parameters.

Although PVT can serve as an alternative to CNN backbones (\eg, ResNet, ResNeXt), there are still some specific modules and operations designed for CNNs and not considered in this work, such as 
SE~\cite{hu2018squeeze}, SK~\cite{li2019selective}, dilated convolution~\cite{yu2015multi}, model pruning~\cite{han2015deep}, and NAS~\cite{tan2019efficientnet}.
Moreover, with years of rapid developments, there have been many well-engineered CNN backbones such as Res2Net~\cite{gao2019res2net}, EfficientNet~\cite{tan2019efficientnet}, and ResNeSt~\cite{zhang2020resnest}.
In contrast, the Transformer-based model in computer vision is still in its early stage of development. 
Therefore, 
we believe there are many potential technologies and applications (\eg, OCR~\cite{wang2019shape,wang2020ae,wang2021pan++}, 3D~\cite{hui2020progressive,cheng2021sspc,hui2021efficient} and medical~\cite{fan2021concealed,fan2020pranet,ji2021progressively} image analysis) to be explored in the future, and hope that PVT could serve as a good starting point.

\section*{Acknowledgments}
This work was supported by the Natural Science Foundation of China under Grant 61672273 and Grant 61832008, the Science Foundation for Distinguished Young Scholars of Jiangsu under Grant BK20160021, Postdoctoral Innovative Talent Support Program of China under Grant BX20200168, 2020M681608, the General Research Fund of Hong Kong No. 27208720.

{\small
\bibliographystyle{ieee_fullname}
\bibliography{egbib}
}

\end{document}